\pdfoutput=1
\documentclass[10pt,twocolumn,letterpaper]{article}

\usepackage{cvpr}              

\usepackage{graphicx}
\usepackage{amsmath}
\usepackage{amssymb}
\usepackage{booktabs}

\usepackage[pagebackref,breaklinks,colorlinks]{hyperref}

\usepackage{graphicx}
\usepackage{multirow}
\usepackage{booktabs}
\usepackage{enumitem}
\usepackage{color}

\usepackage{float}
\usepackage{wrapfig}
\usepackage{tabularx}
\usepackage{mathbbol}
\usepackage{amsmath}
\usepackage{amssymb}
\usepackage[table]{xcolor}
\usepackage{colortbl}

\newcommand{\tablestyle}[2]{\setlength{\tabcolsep}{#1}\renewcommand{\arraystretch}{#2}\centering\footnotesize}

\usepackage[capitalize]{cleveref}
\crefname{section}{Sec.}{Secs.}
\Crefname{section}{Section}{Sections}
\Crefname{table}{Table}{Tables}
\crefname{table}{Tab.}{Tabs.}

\def\cvprPaperID{6788} 
\def\confName{CVPR}
\def\confYear{2024}

\begin{document}
\title{Separate and Conquer: Decoupling Co-occurrence via Decomposition and Representation for Weakly Supervised Semantic Segmentation}

\author{
Zhiwei Yang$^{1,2,3}$\qquad Kexue Fu$^{4}$\qquad\\
Minghong Duan$^{2,3}$\qquad Linhao Qu$^{2,3}$\qquad Shuo Wang$^{2,3\thanks{Corresponding author.}}$\qquad Zhijian Song$^{1,2,3\footnotemark[1]}$ \\
$^{1}$Academy for Engineering and Technology, Fudan University\\
$^{2}$Digital Medical Research Center, School of Basic Medical Sciences, Fudan University\\
$^{3}$Shanghai Key Laboratory of Medical Image Computing and Computer Assisted Intervention\\
$^{4}$Shandong Computer Science Center (National Supercomputer Center in Jinan)
}

\maketitle
\begin{abstract}
 Weakly supervised semantic segmentation (WSSS) with image-level labels aims to achieve segmentation tasks without dense annotations. However, attributed to the frequent coupling of co-occurring objects and the limited supervision from image-level labels, the challenging co-occurrence problem is widely present and leads to false activation of objects in WSSS. In this work, we devise a '\underline{Se}parate and \underline{Co}nquer' scheme SeCo to tackle this issue from dimensions of image space and feature space. In the image space, we propose to 'separate' the co-occurring objects with image decomposition by subdividing images into patches. Importantly, we assign each patch a category tag from Class Activation Maps (CAMs), which spatially helps remove the co-context bias and guide the subsequent representation. In the feature space, we propose to 'conquer' the false activation by enhancing semantic representation with multi-granularity knowledge contrast. To this end, a dual-teacher-single-student architecture is designed and tag-guided contrast is conducted, which guarantee the correctness of knowledge and further facilitate the discrepancy among co-contexts. We streamline the multi-staged WSSS pipeline end-to-end and tackle this issue without external supervision. Extensive experiments are conducted, validating the efficiency of our method and the superiority over previous single-staged and even multi-staged competitors on PASCAL VOC and MS COCO. Code is available \href{https://github.com/zwyang6/SeCo.git}{here}.
\end{abstract}

\vspace{-1.em}
\section{Introduction}

Weakly supervised semantic segmentation (WSSS) as an annotation-efficient alternative to fully supervised semantic segmentation, has enjoyed enormous popularity in recent years~\cite{0}. It aims to densely classify every pixel of an input image by only leveraging more accessible labels than pixel-wise labels, such as points~\cite{1}, scribbles~\cite{2,3}, bounding boxes~\cite{4,5}, and image-level labels~\cite{6,7}. Among these forms of annotations, image-level labels are the most economical yet challenging annotation form to accomplish the segmentation task, as they only indicate the presence of objects and contain the least semantic information.
\begin{figure}[t!]
  \centering
  \includegraphics[width=8.36cm]{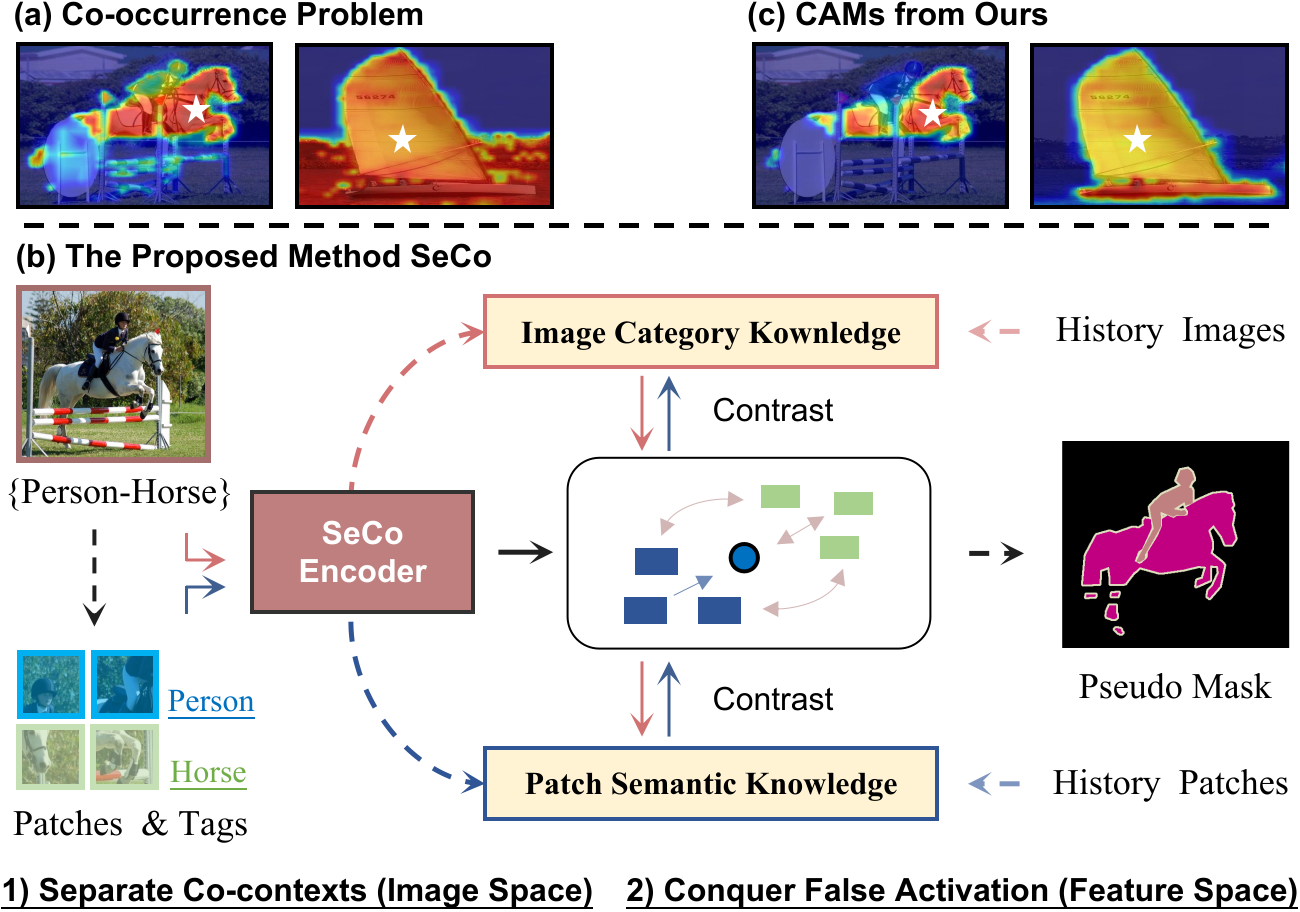}
   \caption{(a) Co-occurrence issue. Targets marked by stars (horse and boat) are falsely activated. (b) To solve this issue, we propose a single-staged framework SeCo, which acts in a 'separate and conquer' manner that efficiently tackles co-occurrence issue without external supervision. It initially separates spatial con-texts in the image space and then conquers false activation in feature space. (c) The proposed SeCo accurately localizes the co-categories.}
   \label{fig.1}
   \vspace{-1.5em}
\end{figure}

Formally, the pipeline for WSSS with image-level labels consists of three steps, i.e., initially training a classification model to generate CAM seeds~\cite{8}, then refining CAMs to generate pseudo labels, and finally training a segmentation network with the pseudo labels and taking it as the final model to inference~\cite{25,53}. Such a WSSS paradigm can be further divided into multi-staged and single-staged groups. For multi-staged WSSS methods~\cite{23,11,13}, the classification and segmentation models need to be trained progressively. It intends to have better segmentation performance while more complicated at training streamline. The single-staged methods~\cite{25,26,27,43} share the encoder for classification and segmentation networks, thus can be trained end-to-end. It is more efficient to optimize while holding inferior performance to the multi-staged. In this study, we focus on the most challenging WSSS paradigm with image-level annotations and streamline the paradigm end-to-end.

For both single- and multi-staged WSSS, generating reliable CAMs from image-level labels is the first and fundamental step for the performance~\cite{9,10}. However, since objects intend to co-occur together, such as \{train, railroad\}, \{boat, water\}, \{horse, person\}, etc., it is inevitable to tackle the co-occurrence of objectives in WSSS. The challenging co-occurrence problem is widespread and often leads to false activation~\cite{t1,t2,11,14}, as shown in \cref{fig.1} (a). Although most existing works have succeeded in completing CAMs, they barely pay attention to such issue, consequently limited to tackling the false activation and bottlenecked in WSSS performance. Recently, introducing external supervision or human prior is proposed to tackle this problem. ~\cite{11,13} leverage the vision-language matched CLIP model~\cite{12} to help distinguish among coupled contexts.~\cite{14} elaborately applies hard out-of-distribution samples to suppress spurious background cues. \cite{15} constructs additional co-categories to address this issue. Although impressive, they heavily rely on external data or elaborate designs to tackle co-occurrence, which impedes real-world applications with complex relations among categories. 

Essentially, the co-occurrence problem arises because the co-appeared contexts coupled in images confuse the networks and incur wrong semantic bias during the feature representation, resulting in false positive pixels activated with high probability. Previous methods ignore the importance of separating co-contexts before representation, thus requiring external data or designs to tackle this issue. Based on the analysis above, we argue that \textbf{initially separating the coupled objects to remove the bias and then enhancing category-specific representation provides a potential insight} to address this issue without external supervision.

As illustrated in~\cref{fig.1} (b), we propose a single-staged WSSS framework SeCo that does not require any extra supervision to tackle the problem. \textbf{Our key insight lies in the 'separate and conquer' training scheme} that decouples co-occurrence in image space and feature space by image decomposition and representation enhancement, respectively. \textbf{(1)} \textbf{To separate the spatial dependence of co-contexts in image space}, we propose to decompose the integral image into multiple patches containing single category information. Previous patch-based method~\cite{51} simply gives image-level labels to patches, but the image labels cannot help differentiate the co-contexts at patch level (see \cref{sec2.1}). Instead, we further explore the strategy to spatially separate coupled objects and focus on tagging each patch. Specifically, category tags from CAMs are designed for each patch, which help identify the co-contexts at patch level. Since CAMs inevitably bring noise in tags, a similarity-based rectification method is designed to revise the noisy tags. In addition, a tag memory pool is constructed to store all history tags, guiding the subsequent patch representation. \textbf{(2)} \textbf{To conquer the co-context confusion and enhance the semantic representation in feature space}, we design a dual-teacher single-student architecture to promote the discrepancy among co-categories. We first build a global teacher to extract category knowledge from integral images. The knowledge provides class centroids for the student in patch representation and helps to push apart co-contexts. Considering the trade-offs between the separation of co-categories and the destruction on global contexts, we share the encoders of both branches to provide complementary information for patch and image semantics. In addition, patch semantic knowledge is further extracted from a patch-level semantic reservoir maintained by a local teacher. Guided by the tags from memory pool, the knowledge helps remove the bias during the representation and pushes apart co-contexts while pulling together those within the same category at a fine-grained level. \textbf{(3)} \textbf{Along with the category tags and the extracted knowledge, multi-granularity contrast is further proposed} across the whole dataset to decouple the co-contexts deeply.

Extensive experiments are conducted on PASCAL VOC and MS COCO, validating its effectiveness in tackling co-occurrence (as shown in \cref{fig.1} (c)) and the superiority over previous single-staged and multi-staged competitors.

\section{Related Works}
\begin{figure*}[t!]
  \centering    
  \includegraphics[width=17.2cm]{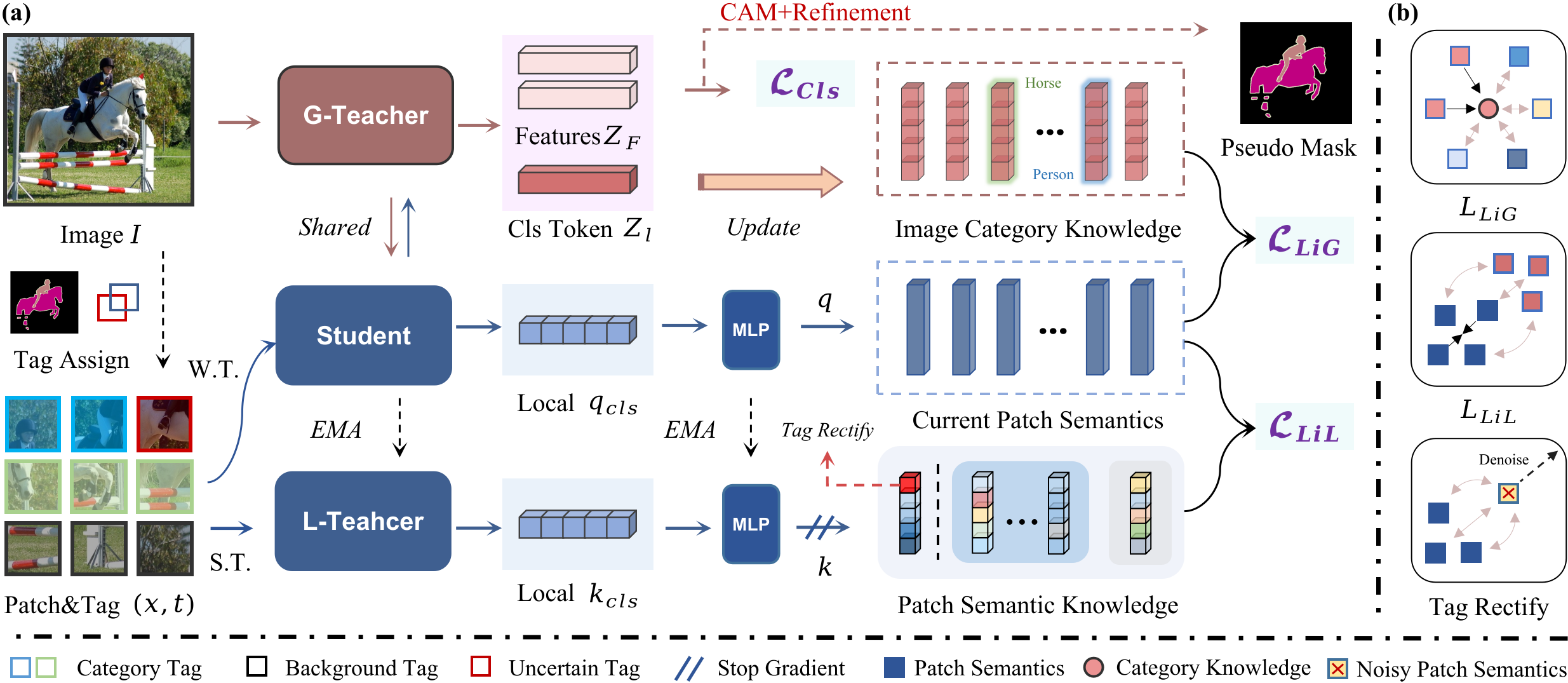}
   \caption{(a) Architecture of the proposed SeCo to tackle co-occurrence issue. Specifically, integral images are firstly sent to the global teacher (G-Teacher) to extract the category knowledge and CAMs. Then three types of category tags, i.e., single-category, background and uncertain tags, are generated from CAMs and allocated to patches accordingly. With tags, two views of patches by different augmentations, i.e., weak data augmentation (W.T.) and strong augmentation (S.T.), are sent to the student and local teacher (L-Teacher) branch, respectively. The local teacher stores all the history patches and category tags and generalizes the patch semantic knowledge. Finally, two contrastive losses, ${L}_{{LiG}}$ and ${L}_{{LiL}}$, are conducted to guarantee the decoupling. In addition, CAMs from global teacher are refined as pseudo labels to train the segmentation network. Since the encoders of segmentation model and classification model are shared, our WSSS pipeline can be trained end-to-end. (b) Illustrated essence of the key components in SeCo. More details are introduced in \cref{sec.3.1}}
   \label{fig.2}
   \vspace{-1.5em}
\end{figure*}
\subsection{Learning from Local Semantics}
\label{sec2.1}
The image-level labels provide limited supervision to generate high-quality CAMs, which motivates many researches to dig complementary information from local semantics. Decomposing a whole image into local patches offers a practical implementation. L2G~\cite{51} crops patches to mine different views of semantics and generates more complete CAMs. PPL~\cite{t3} utilizes feature patches to explore scattered local details and generates CAMs to cover the whole object. ToCo~\cite{27} extracts local semantics from patches and activates more non-discriminative areas. However, simply giving image-level labels to each patch like L2G cannot identify co-categories at patch level, or grouping them into foregrounds and backgrounds as ToCo fails to tell co-categories in the foregrounds. Both of them aggravate the unexpected co-occurring bias during patch representation and suffer from false activation. To this end, we observe that rich semantics from CAMs can be potentially used to tag each patch, which helps differentiate co-contexts and guide category representation at a patch level. To the best of our knowledge, we are the first to report that assigning class-specific tags to patches effectively separates co-categories and reduces the bias during the representation. 

\subsection{Contrastive Learning \& Knowledge Distillation}
Contrastive learning (CL)~\cite{28,29,t6} and knowledge distillation (KD)~\cite{49,50} are two prevalent techniques to promote feature representation. Inspired by it, RCA~\cite{48} and ToCo~\cite{27} conduct region-level contrast to focus on non-discriminative pixels. PPC~\cite{t4} leverages prototype-based~\cite{t5} contrast to expand CAMs. Apart from incorporating CL, SCD~\cite{52} and L2G~\cite{51} introduce knowledge distillation into WSSS and succeed in completing CAMs. However, since the co-occurring pixels intend to be falsely activated with high confidence, simply conducting CL or KD without separating co-contexts could bring much noise to the feature representation, consequently limited to suppressing false activation and tackling co-occurrence. Different from previous works, we focus on reducing noise during feature representation from a new perspective of a 'separate and conquer' paradigm with those techniques, and highlight removing the dependence among co-contexts from two dimensions of image space and feature space, respectively.

\section{Methodology}

\subsection{Problem Definition}
The co-occurrence problem stems from the fact that the coupling of co-contexts confuses networks and introduces noise during feature representation, leading to false positive pixels still activated with high confidence. Therefore, the key of SeCo to addressing the problem is to initially separate the co-occurring objects before feature extraction and then enhance category-specific representations to promote the discrepancy among co-contexts.

\subsection{Framework Overview}
\label{sec.3.1}
\cref{fig.2} (a) specifically depicts the pipeline of SeCo. Given a space of input images ${X}$ and a space of classification labels $\mathcal{Y}=\{1,2, \ldots, K\}$, where the number of categories is denoted as $K$, the training dataset is defined as ${D}=$ $\left\{\left({I}_{{i}},{Y}_{{i}}\right)\right\}_{{i}=1}^{{V}}$. Each tuple $({I}, {Y})$ in ${D}$ is the input, where ${I} \in \mathbb{R}^{3 \times H \times W}$ is the image and ${Y} \in \mathcal{Y}$ is the class label. In our framework, we decompose an image $I$ into patches $x$ and remove the spatial dependence among co-categories. Then we build a dual-teacher-single-student architecture to extract multi-granularity knowledge and conduct semantic contrast to facilitate discrepancy among co-contexts. 

\cref{fig.2} (b) illustrates the essence of the proposed losses ${L}_{{LiG}}$, ${L}_{{LiL}}$ and tag rectification strategy. ${L}_{{LiG}}$ means the loss between local patches and global images. The category knowledge (pink circle) from images acts in centroids for the patch representation (colored squares). ${L}_{{LiL}}$ means the loss among local patches. It pushes apart patch semantics in different categories and pulls together those within the same. The rectification strategy excludes noisy patches in abnormal similarity with the help from category tags.

\subsection{Image Decomposition}
\label{sec.3.2}
\textbf{Assignment of Category Tags.} In order to remove the spatial dependence on co-contexts, we propose to separate co-categories with image decomposition. As shown in \cref{fig.2} (a), given $(I,Y)$ as input, we decompose the integral image $I$ into multiple patches ${x}=\left\{{x}_{{i}} \in \mathbb{R}^{3 \times h \times w}\right\}_{i=1}^{n}$ and the cropping process can be denoted as ${x}={crp}({I})$, where ${crp}(\cdot)$ is the cropping operation, ${h} \times {w}$ is the size of each patch, and ${n}$ represents the number of local patches.

The essence of contrastive learning is the construction of positive pairs~\cite{28}. Simply cropping images into patches cannot help differentiate co-categories~\cite{51,27}. To conduct contrast among co-categories, we assign a category tag $t_{i}$ from the raw CAM seeds to each patch ${x}_{i}$ and leverage tags to guide the subsequent class-specific contrast. Specifically, samples with the same tags are viewed as positive pairs, while samples with the different are negative pairs. We firstly generate CAM seeds by introducing an auxiliary classification head in the teacher network. It is found that the auxiliary head from intermediate features helps generate more diverse CAMs than that from final features~\cite{27}. Although CAMs cannot provide precise supervision, this raw signal effectively guides the assignment of category tags. We obtain the auxiliary pseudo mask ${M}_{aux}$ from ${CAM}_{aux}$:
\begin{equation}                            
    C A M_{aux}={Relu}\left(W_{\lambda}^{T} Z_{F}^{\lambda}\right),
    \label{eq:1}
\end{equation}
where $Z_{F}^{\lambda}$ represents the features from the intermediate ${\lambda}$-th layer of the teacher encoder, ${W}_{\lambda}$ is the mapping matrix in the corresponding classification head, and $Relu(\cdot)$ is the activation function. With $C A M_{aux}$, we obtain the auxiliary pseudo mask ${M_{aux}}$ to guide the allocation of category tags $t = \{t_{i}\}_{i=1}^{n}$ for $x$. We have ${m}=crp({M}_{aux})$ and ${m}=\left\{{m}_{{i}} \in \mathbb{R}^{{h} \times {w}}\right\}_{i=1}^{{n}}$, where ${m}_{{i}}$ represents the pseudo mask patch for the image patch $x_{i}$. 

As shown in \cref{fig.2} (a), we divide patches into background type ${t}_{{i}}=0$, single category type $t_{i}=y_{i}$, and uncertain type $t_{i}=-1$, and allocate category tags accordingly. Specifically, a proportion threshold $\varphi$ determines the tag types based on the proportion of target pixels to $m_{i}$. The uncertain tags represent noisy cases of separating co-occurrence and are excluded from the subsequent contrast.

However, since CAM seeds inevitably bring noise, the assigned category tags are possibly incorrect when guiding the contrast among co-contexts. Therefore, we design a rectification strategy to revise the noisy tags and guarantee the contrastive representation. It is detailed in \cref{sec.3.4}.

\textbf{Representation of Local Patches.} Following the setups of popular contrastive approaches~\cite{29}, we generate two augmented views from local patches ${x}$ by implementing weak data augmentation (W.T.) $Aug_{{q}}(\cdot)$ and strong augmentation (S.T.) $Aug_{{k}}(\cdot)$, and send them to the student encoder $g_{q}(\cdot)$ and local teacher encoder $g_{k}(\cdot)$ to extract class embedding ${q_{cls}}$ and ${k_{cls}}$, respectively, as shown in \cref{fig.2} (a). Class token in ViT is used to represent the embedding as it generalizes high-level semantics~\cite{50}. We further adopt a MLP operation ${O}_{{q}}(\cdot)$ and ${O}_{{k}}(\cdot)$ on the obtained embeddings to strengthen the feature and obtain the final representation ${q}=\left\{{q}_{{i}} \in \mathbb{R}^{1 \times {C}}\right\}_{{i}=1}^{{n}}$ and ${k}=\left\{{k}_{{i}} \in \mathbb{R}^{1 \times {C}}\right\}_{{i}=1}^{{n}}$, which denotes the local semantics from patches. The patch representation is formulated as:
\begin{equation}
  {q}=O_{{q}}\left({g}_{{q}}\left(\operatorname{Aug}_{{q}}({x})\right)\right), {k}=O_{{k}}\left(g_{{k}}\left(\operatorname{Aug}_{{k}}({x})\right)\right).
    \label{eq:2}
\end{equation}

\subsection{Representation with Category Knowledge}
\label{sec.3.3}

\textbf{Class-specific Knowledge Extraction.} After spatially separating the co-occurring context, we build a global teacher to dynamically extract category knowledge from integral images. Notably, considering the image decomposition may destruct the semantic context of patches while global CAMs lack local details, we train both teacher and student and share the encoders to facilitate the knowledge communication between local and global semantics. 

Previous works~\cite{32,33} extract semantics based on CAM. SeCo extracts category knowledge $P$ by utilizing the virtue of class token in ViT~\cite{50}. It represents the high-level semantics of each category and avoids the noise from false localization of CAMs. In particular, the knowledge set from images consists of $K$ prototypes and each prototype generalizes the corresponding category semantics, i.e.,$P=\left\{{P}_{l} \in \mathbb{R}^{1 \times C}\right\}_{l=1}^{K}$. Given the input image ${I}$ with categories $l$, we extract the class token $Z_{l}$ from the global teacher encoder $f(\cdot)$ to denote the category representation. The process can be formulated as $Z_{l}=f(I)$. 


To reduce the noise from co-occurring objects and comprehensively generalize the knowledge, we propose an adaptive updating strategy to gather all semantics across the dataset. Given the token $Z_{l}$ with multi-class, we calculate the cosine similarity with the corresponding prototypes and leverage the similarity scores after $softmax(\cdot)$ as the weights $W=\left\{{W}_{l} \in \mathbb{R}^{1 \times C}\right\}_{l=1}^{K}$ to estimate the relevance to the corresponding category knowledge. The updating process can be formulated as:
\begin{equation}
P_{l} \leftarrow Norm\left(\eta P_{l}+ W_{l}\cdot(1-\eta) Z_{l}\right).
    \label{eq:33}
\end{equation}

Particularly, based on the prior that class tokens from single-category images are most relevant to the corresponding category semantics, the single class tokens are only used to update the prototypes and $ W_{l} = 1.0 $ at this time.

\textbf{Knowledge Guided Contrast for Co-categories.} With the global category knowledge and patch semantics $q$, we design ${L}_{LiG}$ loss inspired by InfoNCE~\cite{46} to guide the student training. To promote the discrepancy of co-categories, only the co-categories are viewed as negative pairs while the semantics within the same category is positive pairs. Hence, the diversity comparison is held between the filtered local semantics $q_{s}=\{q_{i}\}_{i=1}^{u}$ and the knowledge ${P}_{{s}}$ with the appeared categories, where $u$ is the number of patches. The contrast between patch semantics and category prototypes is achieved by:
\begin{equation}
{L}_{LiG}=-\frac{1}{N_{g}^{+}} \sum_{i=1}^{u} \log \frac{\exp \left(q_{i}^{T} P_{l}^{+} / \tau_{g}\right)}{\sum_{P_{l} \in P_{s}} \exp \left(q_{i}^{T} P_{l} / \tau_{g}\right)},
    \label{eq:4}
\end{equation}
where ${N}_{{g}}^{+}$ counts the number of positive pairs between patches and prototypes, $P_{l}^{+}$ is the positive prototypes within the same category $l$ with $q_{i}$ and $\tau_{g}$ is the temperature factor.

\subsection{Representation with Patch Semantics}
\label{sec.3.4}
\textbf{Local Semantics Extraction.} Following the memory setup in contrastive learning, we store patch semantics across the dataset. However, simply taking two views of a patch as a positive pair is not helpful to learn the difference among co-contexts. We propose a category tag pool to match the memory bank and guide the contrast among co-categories. Both patch semantics and category tags are stored as supportive knowledge to decouple the co-context at patch level. Specifically, we build a local teacher to extract features from history patches and update the reservoir and tag pool chronologically by storing the most recent key semantics ${k}$ and its corresponding tag $t_{i}$ while dequeuing the oldest. Mathematically, given the input $(x, t)$, the current query and key embeddings with tags are denoted as $B_{q}$ and $B_{k}$, the oldest are $B_{-q}$ and $B_{-k}$, respectively. Then the reservoir paired with tags is defined as:
\begin{equation}
R(x,t)=B_{q} \cup B_{k} \cup queue \backslash \{B_{-q} \cup B_{-k}\},
    \label{eq:5}
\end{equation}
where $queue \in \mathbb{R}^{N \times C}$ is the history local semantics paired with tags in the reservoir and $N$ is the reservoir capacity. 

Importantly, we update local teacher from student with EMA to keep the memories consistent for contrast and avoid the dramatic variance between the older memories and the newest in the reservoir~\cite{29}.

\textbf{Rectification of Noisy Category Tags.} The tags from CAM seeds are inevitably noisy and bring noise in contrast. To remedy it, we propose a similarity-based rectification strategy to denoise the tags. Since the similarity between two patches with the same category should be significantly higher than those different~\cite{48}, we leverage the memories in the reservoir to rectify noisy tags in an unsupervised manner. When embedding $k_{i}$ with a tag $t_{i}$ updates the reservoir, we compute the inner product between its query view $q_{i}$ and history embeddings $R(x,t_{i})_{+}$ to measure the similarity. Then the average similarity $\mu\left(q_{i},t_{i}\right)$ is calculated with:
\begin{equation}
\mu\left(q_{i},t_{i}\right)=\frac{1}{\left|R(x,t_{i})_{+}\right|} \sum_{k_{+} \in R(x,t_{i})_{+}}q_{i}^{{T}} k_{+},
    \label{eq:6}
\end{equation}
where $k_{+}$ is the embedding from $R(x,t_{i})_{+}$. Embeddings in the reservoir with noisy category tags hold a smaller fraction, thus the average similarity between the falsely tagged samples and the true positive samples is lower than that between true positive samples. Once the number of abnormal-similarity pairs exceeds a certain proportion $\sigma$, we consider $q_{i}$ as a noisy embedding eventually. At this point, we change the category tags $t_{i}$ to uncertain and exclude them from the contrast. The rectification process is denoted as:
\begin{equation}
t_{i} \leftarrow -1, i f \frac{N_{v}}{\left|R(x,t_{i})_{+}\right|}>\sigma,
    \label{eq:7}
\end{equation}
where $N_{v}=\sum_{k_{+} \in R(x,t_{i})_{+}} \mathbb{1}\left(q_{i}^{T} k_{+}<\mu\left(q_{i},t_{i}\right)\right)$ is the number of noisy pairs.

\textbf{Tag Guided Contrast for Co-categories.} With the category tags and history patch semantics, we design a contrastive loss ${L}_{LiL}$ to differentiate co-categories at patch level. The semantics with the same category tag is viewed as positive pairs and the noisy patch semantics is excluded. With the query semantics $q$ and local embeddings $R(x,t_{i})_{+}$, we apply the loss to supervise the above process. The contrast among patch semantics is denoted as: 
\begin{equation}
{L}_{LiL}=-\frac{1}{N_{l}^{+}} \sum_{i=1}^{n} \sum_{k_{+}} M_{f} \log \frac{\exp \left(q_{i}^{T} k_{+} / \tau_{l}\right)}{\sum_{k^{\prime} \in R(x,t)} \exp \left(q_{i}^{T} k^{\prime} / \tau_{l}\right)},
    \label{eq:8}
\end{equation}
where $M_{f}=\mathbb{1}\left(t_{i} \neq-1\right)$ is the rectification mask to exclude noisy patches, ${N}_{1}^{+}$ is the number of positive pairs, ${n}$ is the number of patches and $\tau_{l}$ is a temperature factor.

\subsection{Training Objectives}
\label{sec.3.5}
As shown in \cref{fig.2} (a), loss functions for SeCo consist of two contrast losses, i.e., ${L}_{LiG}$ and ${L}_{LiL}$, and a classification loss ${L}_{{cls}}$. Apart from it, we also implement an auxiliary classification loss ${L}_{ {cls }}^{ {aux }}$ to supervise the generation of auxiliary pseudo masks and allocate tags to local patches. Both ${L}_{ {cls }}$ and ${L}_{ {cls }}^{ {aux }}$ adopt multi-label soft margin loss. The loss objectives of our SeCo are:
\begin{equation}
{L}_{ {SeCo }}={L}_{ {cls }}+{L}_{ {cls }}^{{aux}}+\alpha {L}_{LiG}+\beta {L}_{LiL}.
    \label{eq:9}
\end{equation}
It is noted that the proposed framework SeCo generates the pseudo masks online and is trained end-to-end to achieve the dense segmentation task. The loss for segmentation adopts cross-entropy loss ${L}_{ {seg }}$. Thus, the overall loss is: ${L}={L}_{ {SeCo }}+\gamma {L}_{ {seg }}$. Following previous approaches~\cite{26,27,45}, we leverage regularization losses to enforce the spatial consistency of CAMs and the predicted masks.

\begin{table*}[t]
  \centering
  \caption{Comparisons with SOTAs in mIoU(\%). $\mathcal{M}$:multi-staged, $\mathcal{S}$:single-staged. $\mathcal{I}$:image labels. $\mathcal{SA}$:saliency maps. $\mathcal{E}$: external data.}
   \vspace{-1em}
  \subfloat[Performance on PASCAL VOC~\cite{34}.]
  {
    \hspace{-3.8em}
    \centering
    \begin{minipage}[b]{0.58\linewidth}{
    \begin{center}
    \tablestyle{3.8pt}{1.0}
    \scalebox{1.0}{
    \begin{tabularx}{0.82\linewidth}{@{}l|c|c|cccc@{}}
    \toprule
    Methods                                             & Type                    & Backbone                                 & CAM                                  & Mask                                 & Val                                  & Test                                 \\ \midrule
    AdvCAM~\cite{9} \tiny CVPR'2021 &                                              & ResNet101                                 & 55.6                                 & 68.0                                 & 68.1                                 & 68.0                                 \\
    GSM~\cite{23} \tiny AAAI'2021  &                                              & ResNet101                                 & -                                 & -                                 & 68.2                                 & 68.5                                 \\
    CDA~\cite{15} \tiny CVPR'2021      &                                              & ResNet38                                 & 58.4                                 & 66.4                                 & 66.1            & 66.8            \\
    W-OoD~\cite{14} \tiny CVPR'2022       &                                              & ResNet38                                 & 59.1                                 &72.1  & 70.7                                 & 70.1                                 \\
    CLIMS~\cite{11} \tiny CVPR'2022    &                                              & ResNet101                                & 56.6                                 & 70.5                                 & 70.4                                 & 70.0                                 \\
    L2G~\cite{51} \tiny CVPR'2022      &                                              & ResNet101                                & -                                    & 71.9                                 & 72.1                                 & 71.7                                 \\
    FPR~\cite{t18} \tiny ICCV'2023   &                                              & ResNet101                                 & 63.8    & 66.4            & 70.3                                 & 70.1                                 \\
    OCR~\cite{19} \tiny CVPR'2023      &\multirow{-8}{*}{$\mathcal{M}$}                           & ResNet38                                 & 61.7                                 & 69.1                                 & 72.7                       & 72.0                        \\ \midrule
    1Stage~\cite{25} \tiny CVPR'2020   &                                              & ResNet38                                 & -                                    & 66.9                                 & 62.7                                 & 64.3                                 \\
    AFA~\cite{26} \tiny CVPR'2022      &                        & MiT-B1                                   & 65.0                                 & 68.7                                 & 66.0                                 & 66.3                                 \\
    ViT-PCM~\cite{40} \tiny ECCV'2022  & $\mathcal{S}$                                            & ViT-B/16                                 & 67.7                                 & 71.4                                 & 70.3                                 & 70.9                                 \\
    ToCo~\cite{27} \tiny CVPR'2023     &                      & ViT-B/16                                 & 71.6                                 & 72.2                                 & 71.1                                 & 72.2                                 \\
    \rowcolor[HTML]{EFEFEF} 
    {\color[HTML]{CB0000} \textbf{SeCo(Ours)}}          & {\color[HTML]{CB0000} } & {\color[HTML]{CB0000} \textbf{ViT-B/16}} & {\color[HTML]{CB0000} \textbf{74.8}} & {\color[HTML]{CB0000} \textbf{76.5}} & {\color[HTML]{CB0000} \textbf{74.0}} & {\color[HTML]{CB0000} \textbf{73.8}} \\ \bottomrule
    \end{tabularx}
    }
    \end{center}
    }
    \end{minipage}
  }
  \subfloat[Performance on MS COCO~\cite{35}.]
  {
    \centering
    \begin{minipage}[b]{0.43\linewidth}{
    \begin{center}
    \tablestyle{4.5pt}{1.0}
    \scalebox{1.0}{
    \begin{tabularx}{\linewidth}{@{}l|c|ccc@{}}
    \toprule
    Methods                                              & \multicolumn{1}{l|}{Type} & Backbone       & Sup.                  & Val                                  \\ \midrule
    EPS~\cite{t2} \tiny CVPR'2021 &                           & ResNet101       &                       & 35.7           \\
    RCA~\cite{48} \tiny CVPR'2022   &                           & ResNet101      &                       & 36.8             \\
    L2G~\cite{51} \tiny CVPR'2022  &                           & ResNet101      & \multirow{-3}{*}{$\mathcal{I+SA}$} & 44.2             \\ \cline{4-5} 
    CDA~\cite{15} \tiny CVPR'2021  &                           & ResNet38       &                       & 33.2             \\
    CLIP-ES~\cite{13} \tiny CVPR'2023  &                           & ResNet101      & \multirow{-2}{*}{$\mathcal{I+E}$} & 45.4                                    \\ \cline{4-5} 
    MCTformer~\cite{10} \tiny CVPR'2022 &                           & ResNet38      &                       & 42.0                                 \\
    FPR~\cite{t18} \tiny ICCV'2023  &        & ResNet101       &    & 43.9                                 \\ 
     OCR~\cite{19} \tiny CVPR'2023  & \multirow{-8}{*}{$\mathcal{M}$}                         & ResNet38      & \multirow{-3}{*}{$\mathcal{I}$}                     & 42.5                                 \\\midrule
    1Stage~\cite{25} \tiny CVPR'2020   &      & ResNet38           &                       & -                                    \\
    SLRNet~\cite{43} \tiny IJCV'2022   &                           & ResNet38           &                       & 35.0                                 \\
    AFA~\cite{26} \tiny CVPR'2022   & $\mathcal{S}$                         & MiT-B          &                       & 38.9                                 \\
    ToCo~\cite{27} \tiny CVPR'2023   &      & ViT-B/16          &                       & 42.3                                 \\
    \rowcolor[HTML]{EFEFEF} 
    {\color[HTML]{CB0000} \textbf{SeCo(Ours)}}               &{\color[HTML]{CB0000}}     & {\color[HTML]{CB0000} \textbf{ViT-B/16}} & \multirow{-5}{*}{$\mathcal{I}$}   & {\color[HTML]{CB0000} \textbf{46.7}} \\ \bottomrule
    \end{tabularx}
    }
    \end{center}
    }
    \end{minipage}
  }
  \label{tab.1}%
   \vspace{-0.5em}
\end{table*}%

\begin{figure*}[t]
  \centering
  \includegraphics[width=17.2cm]{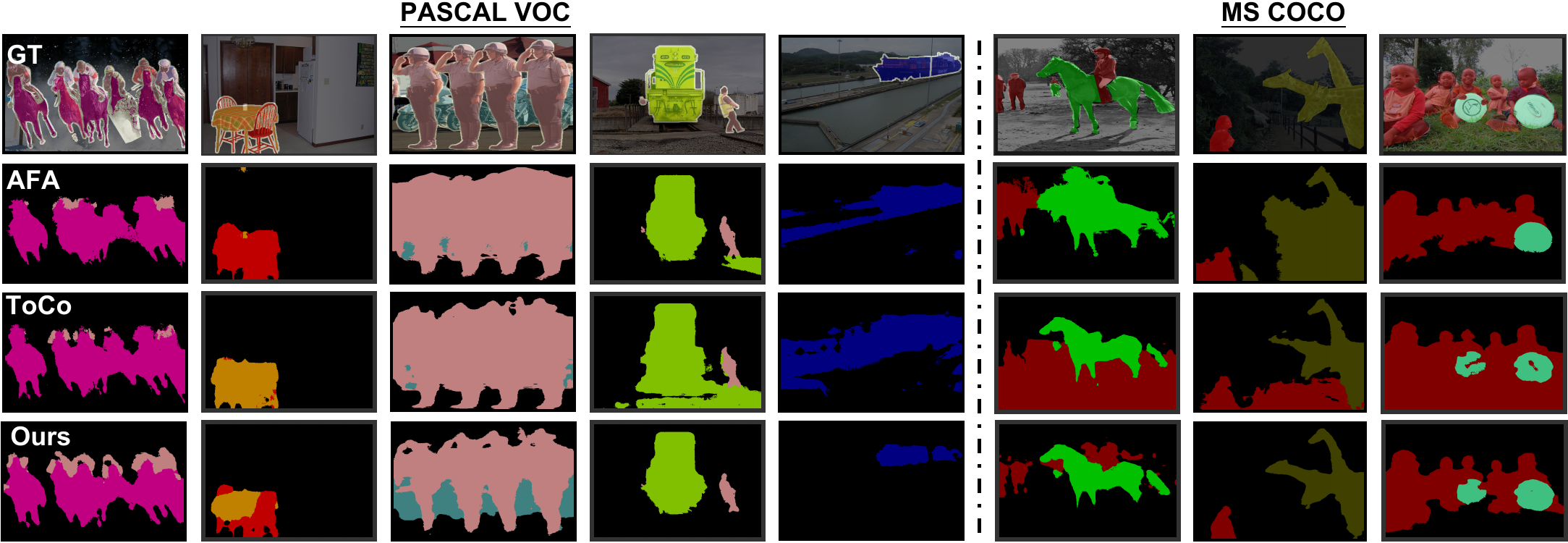}
  \vspace{-1em}
   \caption{Qualitative segmentation results of AFA~\cite{26}, ToCo~\cite{27} and ours on VOC and COCO. SeCo differentiates co-contexts precisely.}
   \label{fig.3}
   \vspace{-1.5em}
\end{figure*}

\section{Experiments and Results}
\subsection{Experimental Settings}

\textbf{Datasets and Evaluation Metrics.} The proposed method is evaluated on PASCAL VOC 2012~\cite{34} and MS COCO 2014~\cite{35}. PASCAL VOC contains 21 semantic categories. Following the practice~\cite{36,26,27}, we use the augmented dataset with $10,582$, $1,449$, and $1,456$ images for training, validating, and testing, respectively. MS COCO includes 81 classes. $82,081$ images are used for training, and $40,137$ images are used for validating. Mean Intersection-Over-Union (mIoU) is used as evaluation criteria. Confusion ratio, i.e., the number of false positive pixels / that of true positives, is designed to evaluate
the efficacy of suppressing false positives from co-occurrence.

\textbf{Implementing Details.} Encoders in the dual-teacher single-student framework all adopt ViT-B~\cite{37} as the backbone and are initialized with pre-trained weights on ImageNet~\cite{38}. Our decoder adopts a simple segmentation head with four $3 \times 3$ convolution layers. Following the training strategy in ~\cite{26,27}, we use AdamW optimizer to train SeCo with a polynomial scheduler. The crop size and the number of the local patches are set to $64\times64$ and $12$. The capacity of the semantic reservoir is $4608$. The loss weight factors $(\alpha, \beta, \gamma)$ in {sec.3.5} are set as $(0.5, 0.5, 0.12)$. All experiments are conducted on RTX 3090 GPU. \textbf{Please refer to Supplementary Materials for more details}.

\subsection{Main Results}
\textbf{Evaluation of CAMs and Pseudo Masks}. \cref{tab.1} (a) quantitatively reports the quality of the initial CAMs and pseudo masks generated by SeCo and other recent competitors on VOC train set. It shows that SeCo generates better CAM seeds with $74.8\%$ mIou, even surpassing other methods refined with post-processing~\cite{44}. With the simple multi-scale refinement~\cite{27}, the quality of pseudo masks further improves to $76.5\%$, significantly higher than both single-staged and multi-staged methods by at least $4.3\%$.

\textbf{Performance of Semantic Segmentation}. \cref{tab.1} (a) reports the performance of the semantic segmentation of SeCo on VOC. The proposed SeCo achieves $74.0\%$ and $73.8\%$ mIoU on the val set and test set, respectively. \cref{tab.1} (b) compares our segmentation performance with other recent methods on COCO val set. Without external data, SeCo achieves $46.7\%$ mIoU and outperforms multi-staged ~\cite{13} competitors tackling co-occurrence and single-staged~\cite{27} SOTAs by $1.3\%$ and $4.4\%$ mIoU, respectively. 

Prediction results on VOC and COCO are visualized in \cref{fig.3}. It illustrates that SeCo can precisely localize the co-occurring objects on both datasets. For example, our method is capable of filtering the distracting backgrounds (water, railroad) from objects, or differentiating the co-occurring foregrounds (horse, person and bicycle), which demonstrates the competence at addressing co-occurrence.


\begin{table}[t!]
\centering
\caption{Abalation study of SeCo on VOC val set.}
   \vspace{-1em}
    \tablestyle{4.3pt}{1}
    \scalebox{1.}
    {
    \footnotesize
    \begin{tabularx}{\linewidth}{@{}l|cccccc@{}}
    \toprule
    Conditions       & \multicolumn{1}{l}{LiG} & LiL & \multicolumn{1}{l}{Tag Rec.} & Recall        & Precision     & \multicolumn{1}{l}{mIoU} \\ \midrule
    Baseline (ViT-B) &                         &     &                              & -             & -             & 54.2                     \\
    w/o LiG          &                         & \pmb{$\checkmark$}   & \pmb{$\checkmark$}                            & 81.1          & 79.2          & 69.1                     \\
    w/o LiL          & \pmb{$\checkmark$}                       &     &                              & 82.7          & 80.9          & 70.3                     \\
    w/o Tag Rec.     & \pmb{$\checkmark$}                       & \pmb{$\checkmark$}   &                              & 83.8          & 82.6          & 72.4                     \\
    
    \rowcolor[HTML]{EFEFEF} 
    {\color[HTML]{CB0000} \textbf{SeCo}}         & {{\color[HTML]{000000} \pmb{$\checkmark$}}}        & {{\color[HTML]{000000} \pmb{$\checkmark$}}} &{{\color[HTML]{000000} \pmb{$\checkmark$}}}   &{\color[HTML]{CB0000} \textbf{85.0}} &{\color[HTML]{CB0000} \textbf{84.0}} &{\color[HTML]{CB0000} \textbf{74.0}}\\   \bottomrule
    \end{tabularx}
    }
   \label{tab.2}
\end{table}



\begin{table}[!t]
    \centering
    \caption{Comparison of IoU and confusion ratio (in the bracket) with recent methods tackling co-occurrence on VOC val set.}
    \vspace{-1em}
    \tablestyle{7.7pt}{1.2}
    \scalebox{1.}
    {
    \footnotesize
    \begin{tabularx}{\linewidth}{@{}l|cccc@{}}
    \toprule
                       &{AFA~\cite{26}}          & {ToCo~\cite{27}}        & \multicolumn{2}{c}{\color[HTML]{000000} \textbf{SeCo(Ours)}} \\ \midrule
    Train w/(Railroad) & 59.6 {(0.63)}             & 58.0 {(0.75)}              & 62.2  &\cellcolor[HTML]{EFEFEF}{\color[HTML]{000000} {\textbf{(0.54)}}}              \\
    Boat w/(Water)     & 64.6 {(0.42})             & 43.6 {(1.11)}              & 68.4 &\cellcolor[HTML]{EFEFEF}{\color[HTML]{000000} {\textbf{(0.32)}}}              \\
    Aeroplane w/(Sky)  & 79.3 {(0.12)}             & 77.3 {(0.19)}              & 86.3 &\cellcolor[HTML]{EFEFEF}{\color[HTML]{000000} {\textbf{(0.07)}}}              \\
    Chair w/(Sofa)    & 29.6 {(1.09)}              & 35.6 {(0.65)}              & 38.3 &\cellcolor[HTML]{EFEFEF}{\color[HTML]{000000} \textbf{(0.48)}}              \\
    Sofa w/(Chair)    &  44.6 {(0.57)}             & 43.8 {(0.77)}              & 57.4 &\cellcolor[HTML]{EFEFEF}{\color[HTML]{000000} { \textbf{(0.35)}}}              \\
    Horse w/(Person)   & 76.0 {(0.14)}             & 83.4 {(0.09)}              & 85.9 &\cellcolor[HTML]{EFEFEF}{\color[HTML]{000000} {\textbf{(0.05)}}}              \\ \midrule
   \textbf{All Categories} & {66.0 {(0.36)}} & {71.1 {(0.32)}} 
   & 74.0 &\cellcolor[HTML]{EFEFEF}{{\color[HTML]{CB0000} {\textbf{(0.23)}}}} \\ \bottomrule
    \end{tabularx}
    }
   \label{tab.5}
\end{table}


\begin{table}[t!]
\setlength{\abovecaptionskip}{0.2cm} 
\setlength{\belowcaptionskip}{-0.15cm}
\centering
\caption{Efficiency performance of SeCo compared to others. The experiment is conducted on PASCAL VOC with RTX 3090.}
\vspace{-0.5em}
\tablestyle{7.2pt}{1.}
\scalebox{1}{
\footnotesize
\begin{tabularx}{\linewidth}{@{}l|ccccc@{}}
\toprule
$\mathcal{M}$                  & CAM & Refine & \multicolumn{1}{c|}{Decoder} & Val                                  & Test                                 \\ \cline{1-1}
CLIMS~\cite{11} & 101 mins       & 332 mins   & \multicolumn{1}{c|}{635 mins}         & 70.4                                 & 70.0                                 \\ \midrule
$\mathcal{S}$                  & \multicolumn{5}{c}{}                                                                                                                              \\ \midrule
AFA~\cite{26}                          & \multicolumn{3}{c|}{554 mins}                                       & 66.0                                 & 66.3                                 \\
ToCo~\cite{27}                           & \multicolumn{3}{c|}{506 mins}                                       & 71.1                                 & 72.2                                 \\

\rowcolor[HTML]{EFEFEF} 
{\color[HTML]{CB0000} \textbf{SeCo(Ours)}}         & \multicolumn{3}{c|}{{\color[HTML]{CB0000} \textbf{417 mins}}}       & {\color[HTML]{CB0000} \textbf{74.0}} & {\color[HTML]{CB0000} \textbf{73.8}} \\ \bottomrule
\end{tabularx}
}
\vspace{-1.5em}
\label{tab.6}
\end{table}

\section{Ablation Study and Further Analysis}

\subsection{Efficacy of Key Components}
Ablative experiments on the key components of SeCo are conducted. \cref{tab.2} shows the segmentation results on VOC val set. Here, w/o LiG means no category prototypes is extracted, w/o LiL means that the local semantic reservoir and category tag pool are not maintained, and w/o Tag Rec. means the tag rectification is not incorporated. The category prototypes from G-teacher act as class centroids for training while the patch knowledge from L-teacher makes the category semantics more compact, which facilitates the difference among co-contexts. As can be seen, without LiG, the precision and recall drop heavily by $4.8\%$ and $3.9\%$, respectively. It verifies that the proposed method can effectively help suppress the false positives and generate more complete masks. Without LiL, the precision, recall and mIoU drop heavily as well. The tag rectification improves precision from $82.6\%$ to $84.0\%$. It works by guaranteeing the right constructions of positive pairs and negative pairs, reducing the noise in the contrastive representation. 

\begin{figure}[t!]
  \centering
  \includegraphics[width=8.4cm]{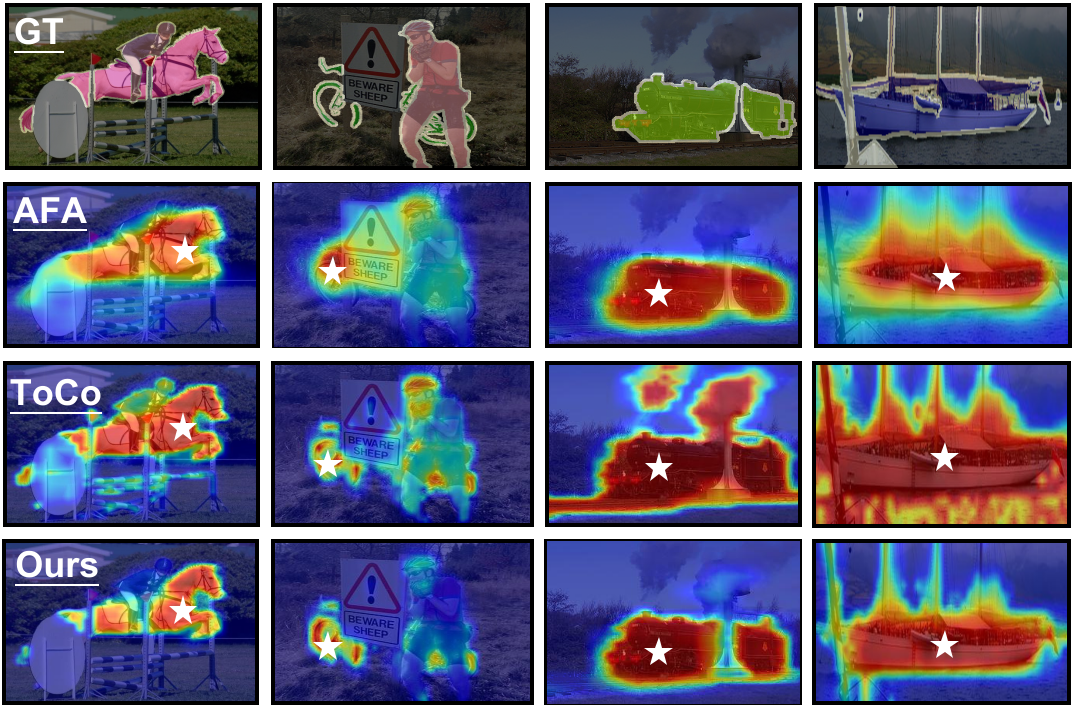}
   \caption{CAMs for co-contexts on VOC between SeCo and competitors~\cite{26,27}. SeCo accurately activates the targets (star).}
   \label{fig.5}
   \vspace{-1.5em}
\end{figure}

\subsection{Effectiveness of Tackling Co-occurrence}
In \cref{tab.5}, representative co-occurring pairs (e.g., \{Boat, water\}, \{train, railroad\}, etc.) in VOC val set are selected to validate the effectiveness of our method to tackle co-occurrence. The metrics of IoU and confusion ratio (in bracket) are adopted. Confusion ratio is calculated by FP/TP, the lower the better. It reports that SeCo demonstrates significantly lower confusion ratio than ToCo, such as boat ({\color[HTML]{CB0000} $-79\%$}), train ({\color[HTML]{CB0000} $-21\%$}), and higher IoU in all representative co-occurring pairs. For all categories, SeCo surpasses other competitors with $0.23$ confusion ratio, lower by at least $9\%$, which shows the superiority of our method to suppress the false positives from co-occurrence.

In addition, we visualize CAMs from SeCo and recent impressive methods in \cref{fig.5}. For the co-occurring objects marked with white stars (horse, bicycle, train, boat), previous methods cannot solve the co-occurrence properly. They are frequently confused by co-contexts and intend to falsely activate the co-occurring backgrounds (water, railroad) or the related foregrounds (person), which can be attributed to the ignoring of noise from co-occurrence during the feature representation. Instead, contributing to the proposed 'separate and conquer' scheme, SeCo shows strength at accurately activating the co-contexts, which could plainly demonstrate the effectiveness in tackling the challenge.

\subsection{Further Analysis}
\textbf{Hyper-parameter Sensitive Analysis.} The analysis of key parameters, such as \textit{patch size, loss weights, reservoir capacity, EMA momentum, temperature factors, etc.}, is specifically discussed in \textbf{Supplementary Materials}. 

\textbf{Training Efficiency Analysis.} 
SeCo is designed in a single-staged paradigm to efficiently tackle the co-occurrence issue. The training efficiency comparisons are reported in \cref{tab.6}. CLIMS~\cite{11} leverages CLIP model to tackle co-occurrence and consists of 3 progressive steps, which takes $1068$ minutes to finish the WSSS workflow. Compared to it, SeCo takes $417$ minutes to finish the workflow and outperforms it by a significant margin. Notably, SeCo achieves more favorable performance compared to other single-staged competitors~\cite{26,27} as well.


\begin{table}[!t]

    \centering
    \caption{The comparison to the fully-supervised counterparts on VOC val set. $\mathcal{I}$:image-level labels. $\mathcal{E}$: external data.}
    \vspace{-1em}
    \tablestyle{7.1pt}{1.1}
    \scalebox{1.}
    {
    \footnotesize
    \begin{tabularx}{\linewidth}{@{}lcccc@{}}
    \toprule
    Methods                                         & Backbone          & Sup.                 & Val                               & Ratio                               \\ \midrule
    DeepLabV2~\cite{t7} \tiny TPAMI'2017 & ResNet101         & \multirow{4}{*}{$\mathcal{F}$}   & 77.7                              & -                                   \\
    WideResNet~\cite{t8} \tiny PR'2019   & ResNet38          &                      & 80.8                              & -                                   \\
    Segfromer~\cite{t9} \tiny ICCV'2021  & MiT-B1            &                      & 78.7                              & -                                   \\
    DeepLabV2~\cite{t7} \tiny TPAMI'2017 & ViT-B/16          &                      & 82.3                              & -                                   \\ \midrule
    \multicolumn{5}{@{}l}{\textbf{Multi-staged methods}}                                                                                                                   \\
    CDA~\cite{15} \tiny CVPR'2021        & ResNet38          & \multirow{3}{*}{$\mathcal{I+E}$} & 66.1                              & 81.8\%                              \\
    W-OoD~\cite{14} \tiny CVPR'2022      & ResNet38          &                      & 70.7                              & 87.5\%                              \\
    CLIMS~\cite{11} \tiny CVPR'2022      & ResNet101         &                      & 69.3                              & 89.2\%                              \\ \cline{3-5} 
    AdvCAM~\cite{9} \tiny CVPR'2021     & ResNet101         & \multirow{3}{*}{$\mathcal{I}$}   & 68.1                              & 87.6\%                              \\
    PPL~\cite{t3} \tiny TMM'2023       & ResNet38          &                      & 67.8                              & 87.3\%                              \\
    MCTformer~\cite{10} \tiny CVPR'2022  & ResNet38          &                      & 71.9                              & 89.0\%                              \\\midrule
    \multicolumn{5}{@{}l}{\textbf{Single-staged methods}}                                                                                                                \\
    1Stage~\cite{25} \tiny CVPR'2020     & ResNet38          & \multirow{6}{*}{$\mathcal{I}$}                  & 62.7          & 77.6\%         \\
    SLRNet~\cite{43} \tiny IJCV'2022        & ResNet38          &   & 69.3          & 85.8\%        \\
    AFA~\cite{26} \tiny CVPR'2022        & MiT-B1            &                      & 66.0         & 83.9\%         \\
    ViT-PCM~\cite{40} \tiny ECCV'2022    & ViT-B/16          &                      & 70.3         & 85.4\%          \\
    ToCo~\cite{27} \tiny CVPR'2023       & ViT-B/16          &                      & 71.1          & 86.4\%          \\
    \rowcolor[HTML]{EFEFEF} 
    {\color[HTML]{CB0000} \textbf{SeCo(Ours)}}         & {\color[HTML]{CB0000} \textbf{ViT-B/16}} &    & {\color[HTML]{CB0000} \textbf{74.0}} & {\color[HTML]{CB0000} \textbf{89.9\%}} \\ \bottomrule
    \end{tabularx}
    }
   \label{tab.7}
   \vspace{-1.5em}
\end{table}


\textbf{Fully-supervised Counterparts.} Since competitors in \cref{tab.1} use different backbones, we report the upper bound performance on VOC val set for fair comparison in \cref{tab.7}. Although ViT intends to have advantageous performance in vision tasks, our method achieves $74.0$ mIoU and $89.9\%$ to its fully-supervised performance, which significantly outperforms other single-staged methods with ViT backbone and holds superiority over other multi-staged competitors~\cite{15,14,11} tackling co-occurrence with external data. 

\begin{figure}[!t]
  \centering
  \includegraphics[width=8.4cm]{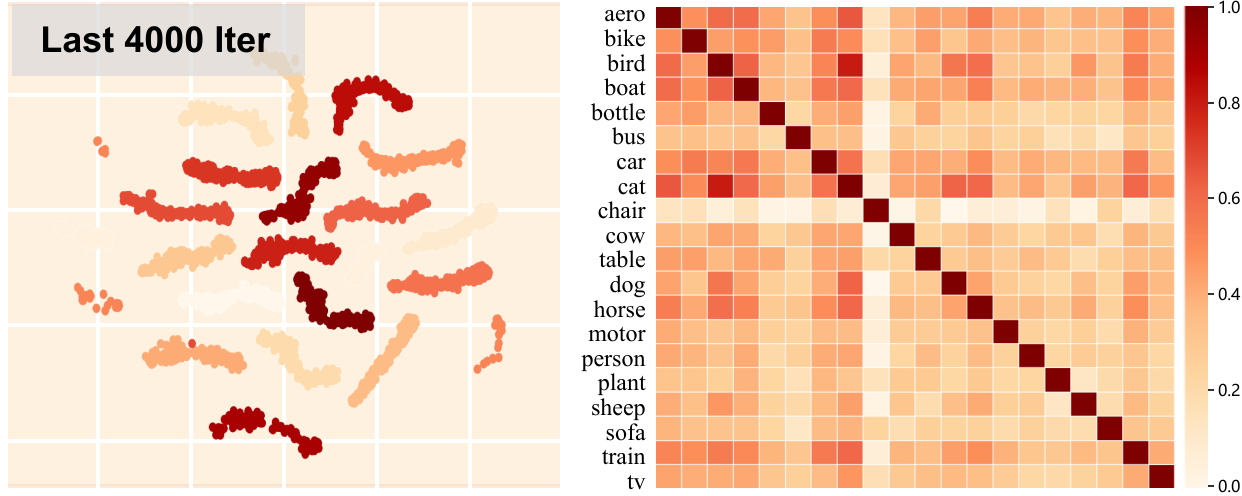}
   \caption{Category representation of SeCo on PASCAL VOC. Left: category prototypes from last $4,000$ iterations are visualized with t-SNE~\cite{t16}. Right: similarity among the category prototypes.}
   \label{fig.6}
   \vspace{-0.6em}
\end{figure}

\textbf{Feature Representation Analysis.} 
We visualize the co-context feature representation at image level to validate the effectiveness. As shown in the right of \cref{fig.6}, we compute the similarity among the category knowledge from images. It is observed that each prototype in the knowledge is only highly related to itself, which suggests that the co-occurring semantics is separated. As shown in the left of \cref{fig.6}, the visualization with t-SNE~\cite{t16} also validates the efficacy. 

\begin{figure}[!t]
  \centering
  \includegraphics[width=8.3cm]{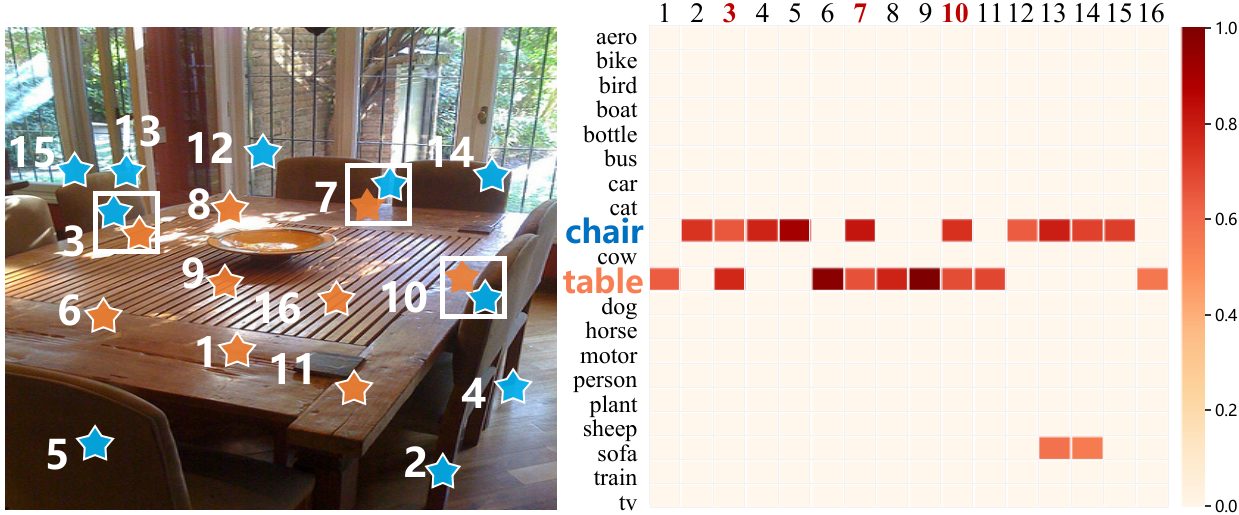}
   \caption{Patch feature representation. Left: sample image with co-contexts \{table, chair\}. The orange star represents the table patch and the blue is the chair patch. Right: similarity among patch semantics and category prototypes. The x-axis means the patch index and the y-axis is the $20$ category prototypes on VOC.}
   \label{fig.7}
   \vspace{-1.3em}
\end{figure}

In addition, we further visualize the co-context feature representation at patch level. We decompose a demo image with co-occurring objects \{dining table, chair\} into $16$ patches, as shown in the left of \cref{fig.7}. Each star (index from $1$ to $16$) denotes a patch, orange stars for table semantics and blue ones for chair semantics. We calculate the similarity between patches and category prototypes. As shown in the right of \cref{fig.7}, the patch semantics has the correct relationship to the corresponding prototypes. Moreover, the patches with index $(3, 7, 10)$ containing both table and chair semantics show a high relationship to both category information. This validates that SeCo can successfully recognize co-contexts instead of being biased to the one or another.

\vspace{-0.2em}
\section{Conclusion}
In this paper, we propose to tackle the widespread co-occurrence problem in WSSS from a new perspective of 'separate and conquer' manner by designing image decomposition and contrastive representation. Extensive experiments are conducted on PASCAL VOC and MS COCO, validating the effectiveness of tackling co-occurrence issue. 

One limitation is that, although we make the attempt to allocate tags and reduce the bias, it inevitably remains co-category patches and allocates wrong tags. In the future, leveraging patches with adaptive size or adopting other denoising techniques is the potential research for WSSS. 
\vspace{-0.4em}
\section{Acknowledgement}
This work was supported by National Natural Science Foundation of China (82072021), Shanghai Municipal Science and Technology Committee (23410710400, 22YF1409300).

{\small
\bibliographystyle{ieee_fullname}
\bibliography{egbib}
}

\end{document}